\documentclass[10pt,twocolumn,letterpaper]{article}

\usepackage{iccv}
\usepackage{times}
\usepackage{epsfig}
\usepackage{graphicx}
\usepackage{amsmath}
\usepackage{amssymb}
\usepackage{subcaption}
\usepackage{subfiles}

\usepackage{booktabs}
\usepackage{color, colortbl}
\usepackage{multirow}
\usepackage{pifont}

\definecolor{Gray}{gray}{0.9}
\definecolor{DarkGray}{gray}{0.5}

\usepackage[pagebackref=true,breaklinks=true,letterpaper=true,colorlinks,bookmarks=false]{hyperref}
\iccvfinalcopy 

\usepackage[capitalize]{cleveref}
\crefname{section}{Sec.}{Secs.}
\Crefname{section}{Section}{Sections}
\Crefname{table}{Table}{Tables}
\crefname{table}{Tab.}{Tabs.}

\def\iccvPaperID{11601} 

\ificcvfinal\fi

\begin{document}

\title{Read-only Prompt Optimization for Vision-Language Few-shot Learning}


\author{Dongjun Lee$^*$ \hspace{0.4cm}  Seokwon Song$^*$ \hspace{0.4cm}  Jihee Suh  \\
Joonmyung Choi \hspace{0.4cm} Sanghyeok Lee \hspace{0.4cm} Hyunwoo J. Kim$^{\dagger}$\vspace{0.4cm}\\
Korea University\\
{\tt\small \{mando03, tjrdnjs99, adelsuh, pizard, cat0626, hyunwoojkim\}@korea.ac.kr}
}

\maketitle
\ificcvfinal\thispagestyle{empty}\fi
\def\thefootnote{*}\footnotetext{Equal contribution.}\def\thefootnote{\arabic{footnote}}
\def\thefootnote{$\dagger$}\footnotetext{Corresponding author.}\def\thefootnote{\arabic{footnote}}

\begin{abstract}
    In recent years, prompt tuning has proven effective in adapting pre-trained vision-language models to downstream tasks. 
    These methods aim to adapt the pre-trained models by introducing learnable prompts while keeping pre-trained weights frozen. 
    However, learnable prompts can affect the internal representation within the self-attention module, which may negatively impact performance variance and generalization, especially in data-deficient settings. 
    To address these issues, we propose a novel approach, Read-only Prompt Optimization (\textbf{RPO}). RPO leverages masked attention to prevent the internal representation shift in the pre-trained model. 
    Further, to facilitate the optimization of RPO, the read-only prompts are initialized based on special tokens of the pre-trained model. 
    Our extensive experiments demonstrate that RPO outperforms CLIP and CoCoOp in base-to-new generalization and domain generalization while displaying better robustness. 
    Also, the proposed method achieves better generalization on extremely data-deficient settings, while improving parameter efficiency and computational overhead.
    Code is available at \url{https://github.com/mlvlab/RPO}.
    
\end{abstract}

\section{Introduction}
\label{sec:intro}

Vision-language models like CLIP~\cite{clip}, ALIGN~\cite{ALIGN}, and FILIP~\cite{FILIP} have achieved excellent performance in various vision-language tasks. Since vision-language models are supervised by natural language based on the contrastive learning objective, by placing the class name in a textual template (e.g.,``A photo of a \texttt{[CLASS]}''), vision-language models can effectively classify images in open-vocabulary settings~\cite{clip}.

\begin{figure}[t]
\centering
{\includegraphics[width=8cm,height=4.5cm]{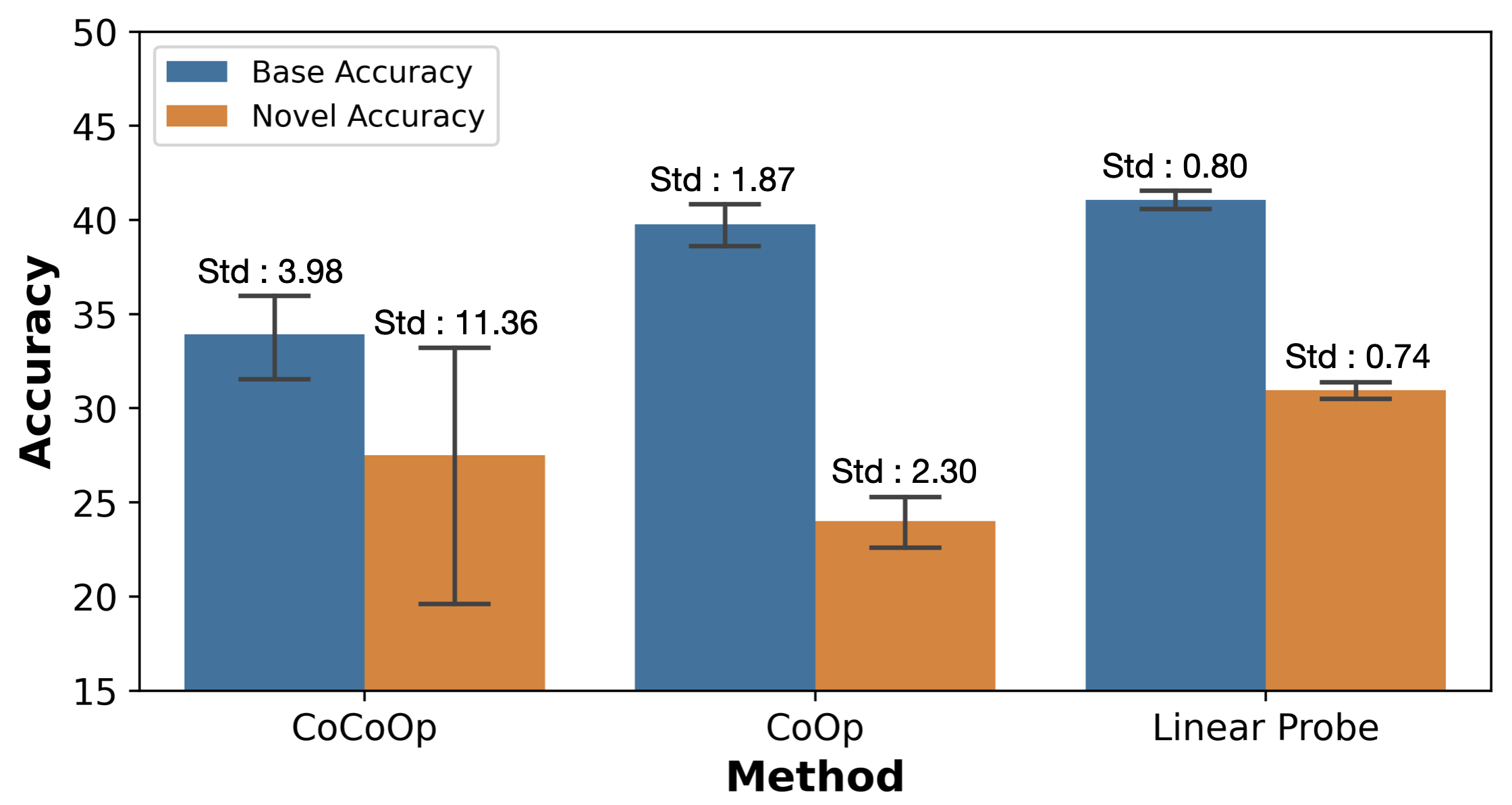}} 
\caption{\textbf{Variance of CoCoOp, CoOp, and linear probing.} Linear probing, which does not shift the pre-trained representation, shows lower variance in performance compared with prompt learning methods such as CoOp and CoCoOp.}
\label{fig:0}
\end{figure}

Recent works have explored the adaptation of these vision-language models on downstream tasks~\cite{clipbenefit}. However, unlike small pre-trained models, large-scale architectures (\textit{e.g.}, CLIP) are difficult to fine-tune, since it is inefficient, resource-intensive, and possibly damaging to the good representations learned during pre-training.
In CLIP, prompt engineering is conducted to provide domain-specific context to downstream tasks (\eg, ``A photo of a \texttt{[CLASS]}, a type of car'')~\cite{clip}. However, this means that the prompt has to be chosen manually, based on trial and error.
 To mitigate this issue, Context Optimization (CoOp)~\cite{coop} suggests automating prompt engineering on CLIP, replacing the context words in natural language-based prompts with learnable vectors. Conditional Context Optimization (CoCoOp)~\cite{cocoop} extended CoOp with an image-conditional prompt, generated by an additional neural network, to improve generalization.

\begin{figure*}
\centering
\subfloat[Conventional Prompt Tuning]{\includegraphics[width=5.3cm,height=7cm]{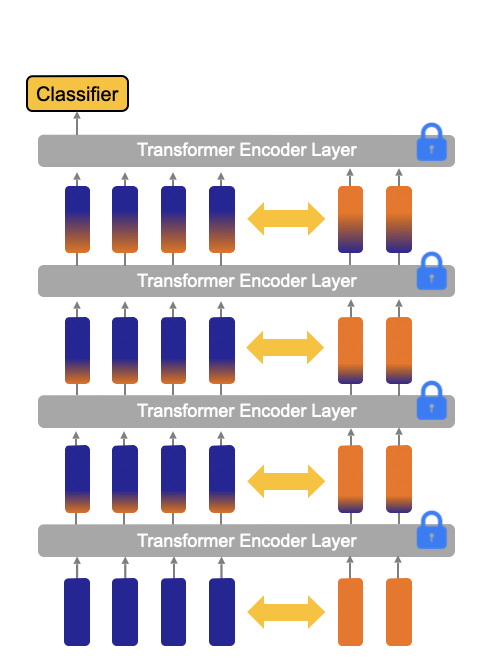}\label{fig:1a}} 
\hspace{0.5cm}
\subfloat[Linear Probing]{\includegraphics[width=5.3cm,height=7cm]{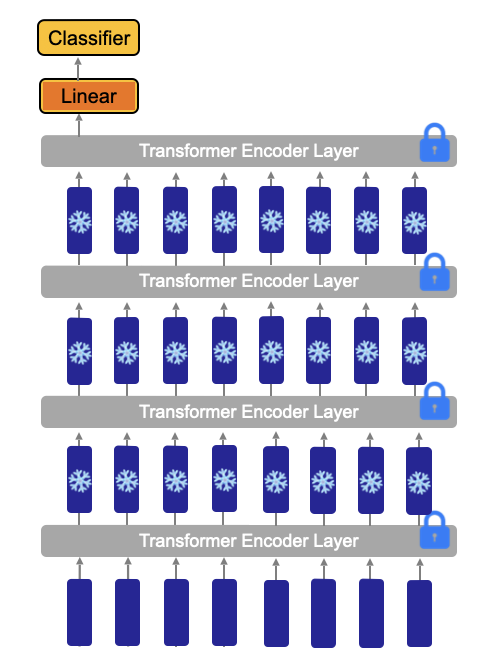}\label{fig:1b}} 
\hspace{0.5cm}
\subfloat[Read-only Prompt Optimization]{\includegraphics[width=5.3cm,height=7cm]{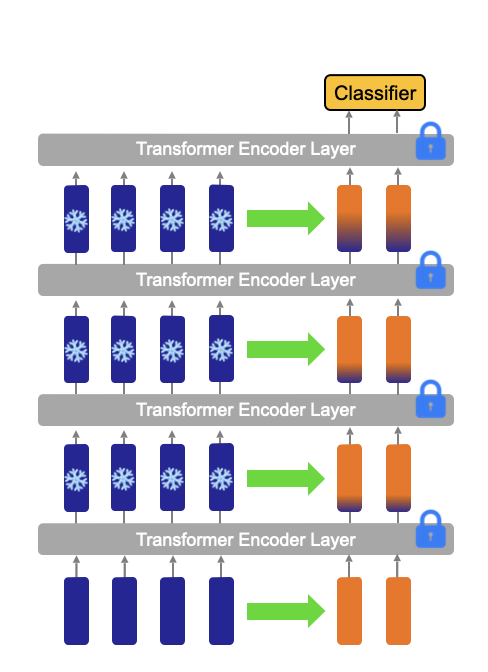}\label{fig:1c}} 
\label{fig:1}
\caption{\textbf{Illustration of methods for model adaptation and RPO.} (a) As denoted by \textcolor[rgb]{0.80,0.8,0}{$\Leftrightarrow$}, token features and prompt features can see each other in conventional prompt tuning methods. Although the weight of the model has been frozen, the internal representations of pre-trained CLIP are increasingly shifted by the newly introduced learnable prompts through the self-attention mechanism. 
(b) In linear probing, internal representations as well as pre-trained parameters are frozen. The linear layer on top of the model is trained for model adaptation.
(c) As denoted by \textcolor[rgb]{0.2,0.8,0}{$\Rightarrow$}, only the prompts can read token features and not the other way around in our method, RPO. This keeps token features frozen and unaffected by introduced prompts while our read-only prompts only read useful information from token features.}%
\end{figure*}

Although these existing methods are proposed to avoid adversely affecting the learned parameters of the pre-trained model during prompt learning, they still affect the model's hidden representation through the attention mechanism, which we call the \textit{internal representation shift}.
We visualize this process of representation shift in \Cref{fig:1a}. As tokens are processed through transformer~\cite{tf} layers, the internal representations of the pre-trained model are largely changed by the learnable prompts. This can be beneficial, as it allows the model to better adapt to the downstream task. However, as shown in \Cref{fig:0}, this shift has the potential to negatively impact the robustness and generalization of the model in data-deficient settings. On the other hand, linear probing has no internal representation shift, as shown in \Cref{fig:1b}, but the linear layer introduces parameter inefficiency.

To inspect how representation shift influences model variance in data-deficient settings, we conduct a preliminary experiment with linear probing CLIP, which does not change the internal representation of pre-trained CLIP. We train the model with 10 random few-shot training data split on the FGVCAircraft dataset with the 16-shot learning setting and visualize the variance of performance. Interestingly, as shown in \Cref{fig:0}, we observed that linear probing significantly lowers variance compared to CoOp and CoCoOp, even though it requires more training parameters (262K) compared to CoOp (2K) and CoCoOp (35K). This result shows that internal representation shifts induced by training with deficient data may result in high variance. At the same time, as CoOp empirically showed, linear probing sometimes shows a lack of generalizability in domain-shift tasks, and the amount of its additional parameters is undesirable.

Motivated by this observation, we propose Read-only Prompt Optimization (RPO) that learns read-only prompts as shown in \Cref{fig:1c}. RPO prevents representation shift during adaptation while being parameter-efficient, leading to a more robust and generalizable adaptation.


Our contributions can be summarized as follows:
\begin{itemize}

  \item {We propose \textbf{Read-only Prompt Optimization} (RPO), which allows prompts only to read information from the attention-based interactions of a pre-trained vision-language model, thereby preventing the internal representation shift.}
  
  \item{We develop a simple yet effective initialization method for our read-only prompts, leveraging the special token embeddings of the pre-trained CLIP vision-language model.}
  \item {Our extensive experiments and analyses demonstrate the generalization of RPO on domain and label shift in few-shot adaptation settings, achieving the best performance in 9 benchmarks on base to new generalization and in 4 benchmarks on domain generalization, at the same time reducing variance depending on the few-shot sample.}

\end{itemize}

\section{Related Works}
\label{sec:relatedwork}

\noindent\textbf{Vision-Language Models} 
The vast amount of web-crawled image-text pairs~\cite{clip, ALIGN, ALT200M, redcaps, wit, laion5b} facilitate vision-language models to be pre-trained contrastively, which enables the acquisition of powerful and generalizable image representations.
For instance, CLIP~\cite{clip} and ALIGN~\cite{ALIGN} rely on transformer-based~\cite{tf} encoders to map the complex relationship between images and text.
These vision-language models have achieved exceptional performance in diverse downstream tasks, especially in zero-shot image classification.  Following these works, numerous other works~\cite{butdatt, biatt, dfvqa, dmvqa} have emerged to harness the power of vision-language models for image-related tasks such as image recognition~\cite{coop, cocoop, vistune, dualprompt, flamingo}. 

However, despite the strong generalization performance of these models, adapting them to specific tasks can be challenging, as assembling large datasets for diverse downstream tasks is a formidable challenge~\cite{clipadapt}. To mitigate this issue, recent works focus on enabling the rapid adaptation of pre-trained vision-language models to specific tasks based on the transferability of CLIP.\\

\noindent\textbf{Prompt Learning}
Prompt learning~\cite{p_tuning, power_of_scale, gpt_understands, ppt} is initially proposed in natural language processing models like GPT~\cite{gpt2, gpt3}, and BERT~\cite{bert}. This technique involves incorporating additional tokens, such as handcrafted instructions or learnable prompts, to facilitate the fine-tuning of a pre-trained language model for downstream tasks. The additional tokens provide contextual information of downstream tasks to the model while keeping the original language model unchanged, thereby avoiding catastrophic forgetting~\cite{catforget}. Based on the effectiveness of this approach, recent studies have tried to utilize the concept of prompt learning in vision-language models. 


Recent studies in vision-language models used prompt learning, with continuous vector prompts which are concatenated and processed with text tokens~\cite{coop, promptvideo}. Another line of works introduced prompts that depend on visual features~\cite{vistune, cocoop, dualprompt, ovsgg, vtclip}. The continuous prompt learning method~\cite{NLP_promptuning_1, NLP_promptuning_2, prefix} reduces the number of parameters to train and automatically identifies a well-functioning prompt. Visual Prompt Tuning (VPT)~\cite{vistune} inserts prompts to the visual encoder rather than the text encoder. Likewise, prompts effectively contain and communicate knowledge about the task at hand.\\ 


\noindent\textbf{Zero-Shot Learning \& Domain Generalization} Zero-shot learning involves learning general knowledge from ``base'' object classes, which is available during training, and using this knowledge to recognize novel classes. To achieve this, some approaches include using visual attributes like color or shape to generalize across classes~\cite{zslAttribute}, or using vision-language models to map visual samples and corresponding text~\cite{coop, cocoop, promptdet}.

Domain generalization requires the visual encoder to generate domain-invariant representations, meaning they are not affected by the particular domain or setting in which the images were taken. For example, a photo of an apple and a sketch of an apple~\cite{imagenetsketch} should result in similar representations. Various methods have been proposed to achieve domain generalization, such as using pre-trained models for generalized representations~\cite{miro, predomgen} and cross-modality supervision~\cite{gvrt}.

While prompt learning in vision-language models has shown improved performance, learnable prompts have a high chance of altering well-functioning parts of the original model through the mechanism of attention~\cite{tf}. The attention mechanism causes all input embeddings to interact with each other, thereby affecting the hidden representation of the pre-trained model. This may lead to unexpected behavior in the frozen model if the training data is insufficient.\\

\newcommand{\mycomment}[1]{}

\section{Method}
\label{sec:method}

\begin{figure*}[ht]
\centering
{\includegraphics[width=17cm, height=7cm]{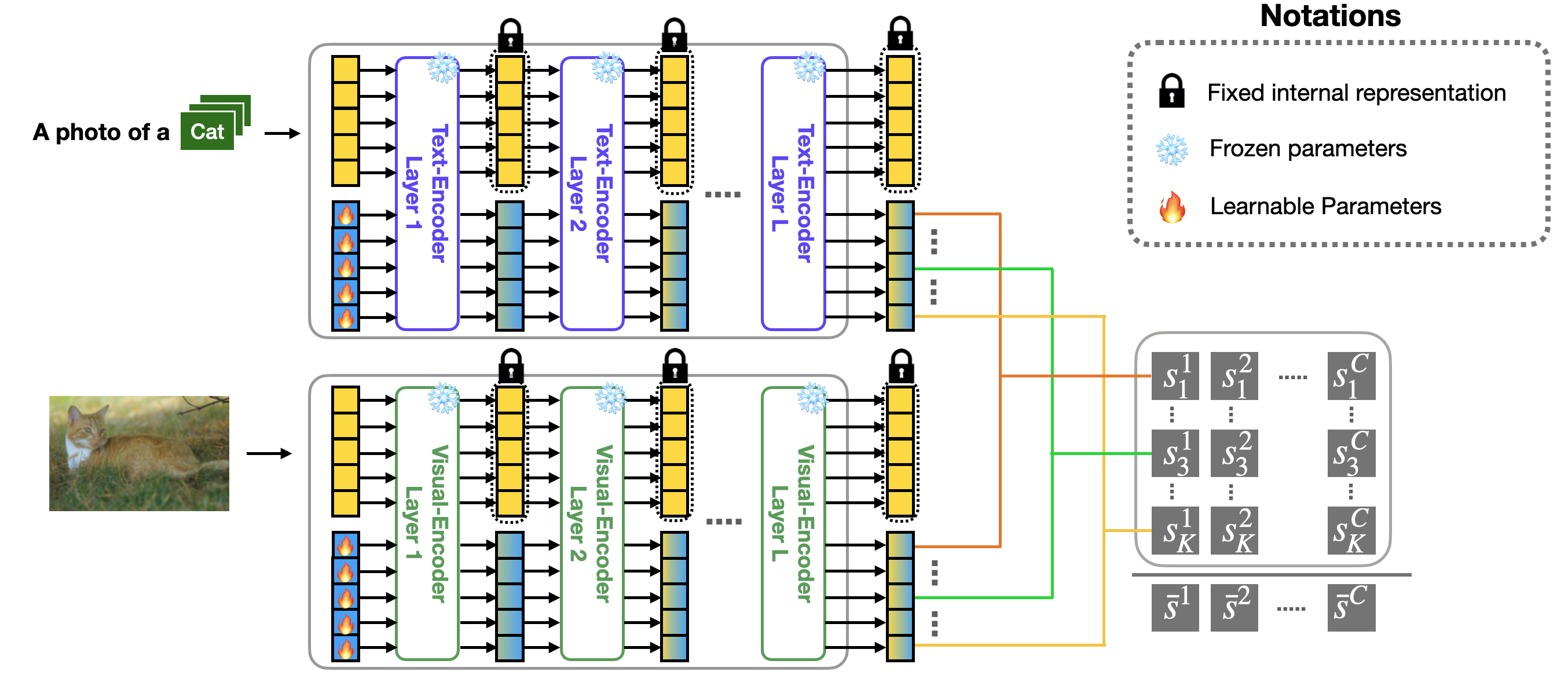}\label{fig:2}}
\caption{\textbf{Overall architecture of RPO.} We use the default prompt ``A photo of a \texttt{[CLASS]}'' for all datasets. Then in both encoders, our read-only prompts are concatenated to the original features and fed into a frozen encoder. Attention within these encoders are masked so that our prompts can be learned, but not shift the original feature interactions. 
We compute similarity scores between the outputs of each encoder corresponding to each of $K$ prompts and average them to produce final classification scores $\bar{s}^1$ to $\bar{s}^C$, where $C$ denotes the number of classes.}
\label{fig:2}
\end{figure*}

In this section, we propose Read-only Prompt Optimization (RPO) for a robust and generalizable adaptation of vision-language models to various downstream tasks in few-shot data deficient settings. We introduce a set of \textbf{Read-only Prompts}, concatenated to the input of the visual and text encoders then processed with \textbf{masked attention} to avoid the impact on the internal representation of CLIP. All pre-trained parameters are frozen during prompt optimization, and only concatenated read-only prompts are updated.

\subsection{Read-only Prompts}
For both the text encoder and visual encoder, RPO works with the same mechanism. We first concatenate a set of continuous learnable prompts, which requires minimal additional parameters to train, to image patch embeddings or word embeddings. The formulation is as below.

\begin{align} 
\mathbf{x^{(0)}} &= \left [x^{(0)} ; E_x^{(0)}; \{p^v_i\}_{i=1}^{K}\right ], \\
\mathbf{y^{(0)}} &= \left [y^{(0)}  ; E_y^{(0)}; \{p^t_i\}_{i=1}^{K} \right], 
\end{align}

where $x^{(0)}\in\mathbb{R}^{d_v}, y^{(0)}\in\mathbb{R}^{d_t}$ denote special token embeddings, \texttt{[CLS]} for the visual encoder and \texttt{[EOS]} for the text encoder, which act as feature aggregators in each encoder. $E^{(0)}_x\in\mathbb{R}^{N_x \times d_v}, E^{(0)}_y\in\mathbb{R}^{N_y \times d_t}$ denote the visual and text embeddings, and $d_v, d_t$ are the dimensions of image patch and word embeddings, while $N_x, N_y$ denote the length of feature tokens, not counting the special tokens.  
$p^v_i, p^t_i$ denotes the $i$th learnable prompt of the visual and text encoder, and $K$ is the number of prompts. The number of prompts is equal for both encoders.
Note that, unlike previous textual prompt learning methods where learnable prompts replace the token embeddings corresponding to `A photo of a', we encode `A photo of a \texttt{[CLASS]}' prompt to produce $E_y^{(0)}$ and then concatenate read-only learnable prompts $\{p_i^t\}_{i=1}^K$. 
\\

\subsection{Special token-based initialization}
In RPO, each learnable prompt is initialized by slightly perturbed special tokens, \textit{i.e.}, \texttt{[CLS]} on the visual encoder and \texttt{[EOS]} on the text encoder, of the pre-trained CLIP, named ST-Initialization.
 In CLIP, special tokens play the role of a feature aggregator which acts as a representative of the input at the last layer of the transformer encoder. Since read-only prompts carry out feature aggregation as well, we discovered that it is beneficial to initialize prompts based on special tokens as a good starting point. The ablation study of ST-Initialization is described in \Cref{tab:3}. We initialize prompts as follows:
\begin{equation} 
\begin{split}
p^v_i \sim \mathcal{N}(x^{(0)},\,\sigma^2I), \quad p^t_i \sim \mathcal{N}(y^{(0)},\,\sigma^2I),
\end{split}
\end{equation}
where $\{{p^v_i\}_{i=1}^{K}} \in \mathbb{R}^{K \times d_v}$ and $\{{p^t_i\}_{i=1}^{K}} \in \mathbb{R}^{K \times d_t}$ denote the set of read-only visual prompts and text prompts, and $\sigma^2$ is the variance for initialization. In this paper, we set $\sigma$ as $0.1$. This initializes $K$ prompts slightly differently so that the learnable prompts avoid constant initialization.\\

\begin{figure}[t]
\centering
\subfloat[Visual Attention Mask]{\includegraphics[width=4.5cm,height=4.7cm]{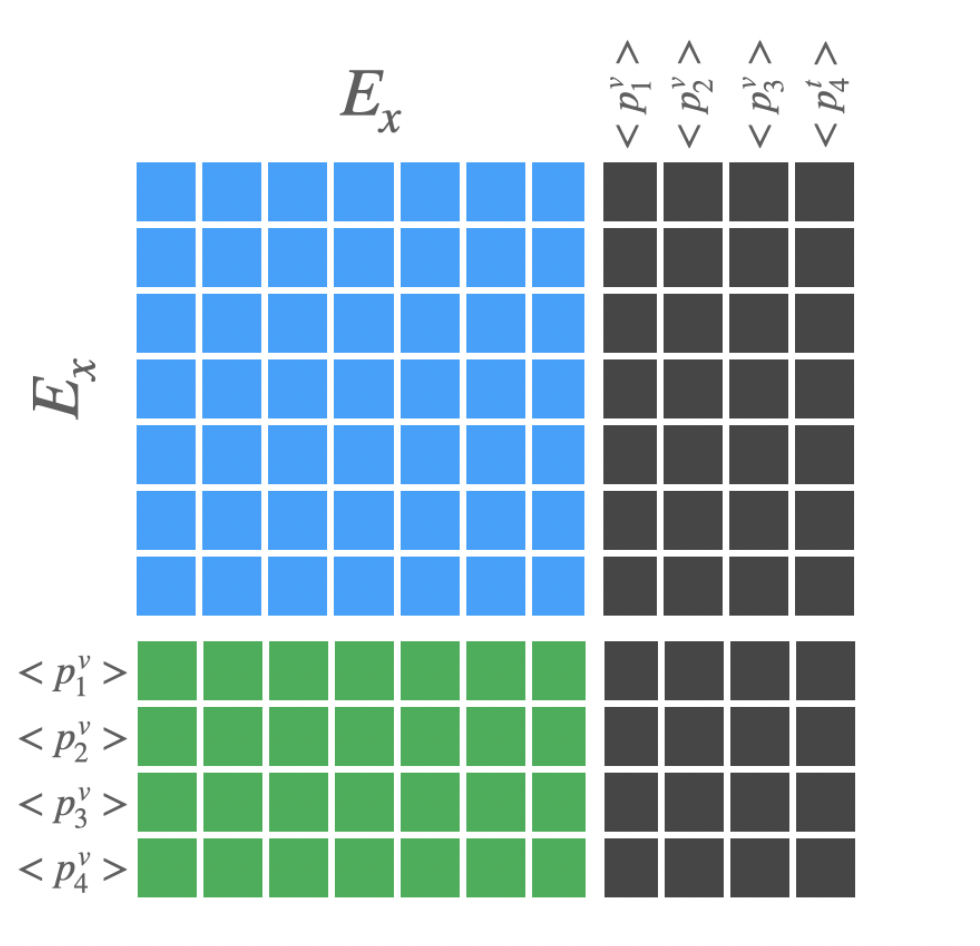}\label{fig:3a}}
\subfloat[Textual Attention Mask]{\includegraphics[width=4.3cm,height=4.7cm]{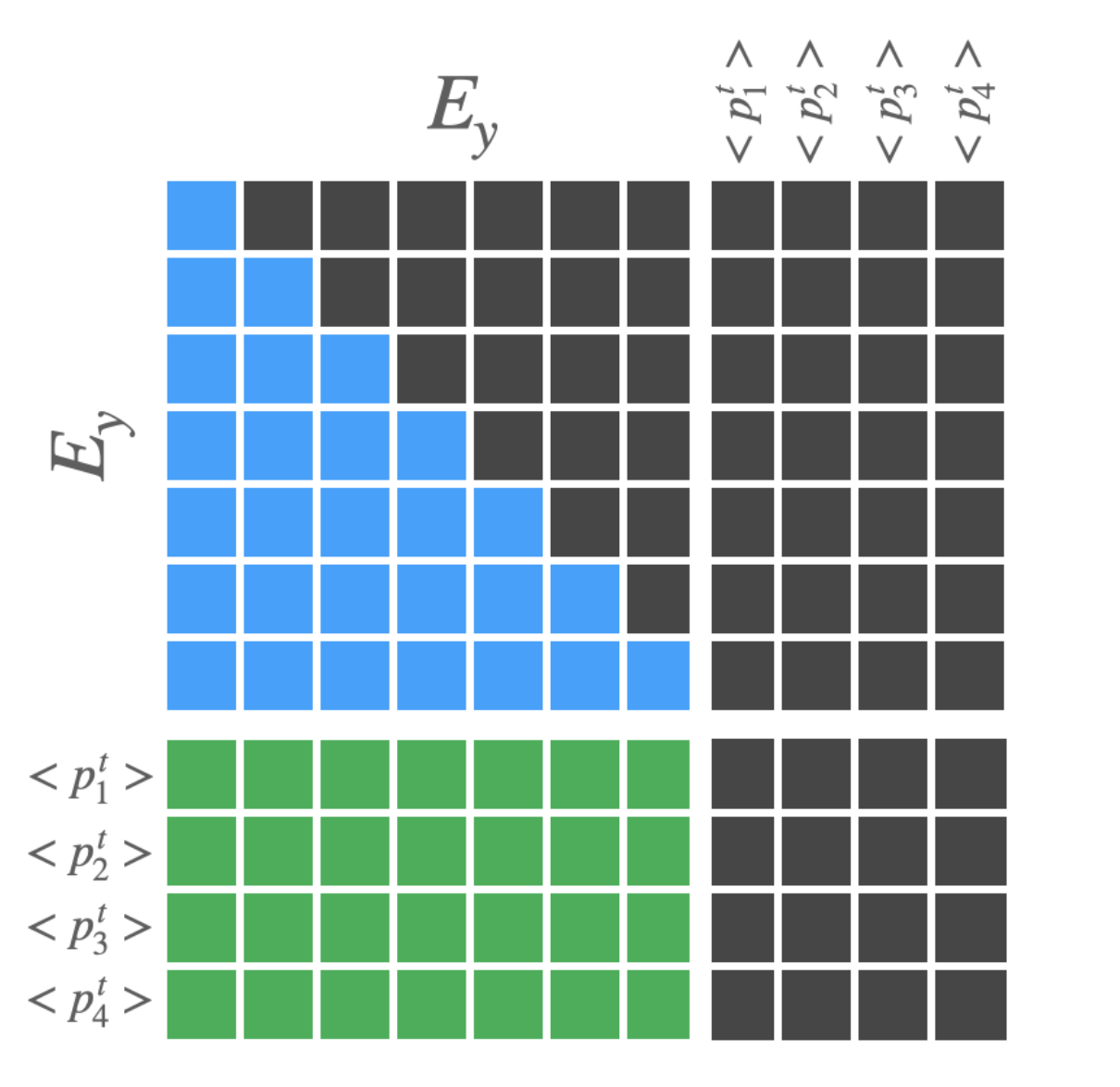}\label{fig:3b}}
\caption{The visualization of attention masks for each encoder. }
\label{fig:3}
\end{figure}

\subsection{Masked attention}

In our framework, RPO, masked attention is important for preserving internal interactions within the pre-trained CLIP. As shown in \Cref{fig:3a} and \Cref{fig:3b}, we propose an attention mask to prevent the original features from being corrupted by learnable prompt embeddings. The visual attention mask $M_v \in \mathbb{R}^{N_v \times N_v}$ and textual attention mask $M_t \in \mathbb{R}^{N_t \times N_t}$ restricts the attention flow from learnable prompts to existing features, where $N_v = 1 + K + N_x$ and $N_t = 1 + K + N_y$.

The mask can be defined as follows, where $M^{i, j}$ denotes the $i$th row, $j$th column element of the mask: 
\begin{flalign}
 M^{i, j}_v &= \begin{cases}
    -\infty, & \text{if } j > 1 + N_x \\
    0,              & \text{otherwise}
\end{cases} \\
M^{i, j}_t &= \begin{cases}
    -\infty, & \text{if } j > 1 + N_y \text{ or } i > j\\
    0,              & \text{otherwise}
\end{cases}
\end{flalign}
Masked attention operations in the transformer encoder can be formulated as below.
\begin{equation}
\begin{split}
\mathbf{x}^{(l+1)} & = \mathcal{V}_{l+1}(\mathbf{x}^{(l)}, \mathit{M_v}) \\
& = \mathbf{softmax}\left(\frac{QK^T}{\sqrt{d_v}} + \mathit{M_v}\right) \cdot V,\\
\mathbf{y}^{(l+1)} & = \mathcal{T}_{l+1}(\mathbf{y}^{(l)}, \mathit{M_t}) \\
& = \mathbf{softmax}\left(\frac{QK^T}{\sqrt{d_t}} + \mathit{M_t}\right) \cdot V,
\end{split}
\end{equation}
where $\mathcal{V}_{l+1}$ and $\mathcal{T}_{l+1}$ are the $(l+1)$-th masked multi-head self-attention layer of the visual encoder and text encoder, respectively. $\mathbf{x}^{(l)}\in\mathbb{R}^{N_v \times d_{v}}$ denotes the input tensor of the $(l+1)$-th visual encoder layer and $\mathbf{y}^{(l)}\in\mathbb{R}^{N_t \times d_{t}}$ denotes the input tensor of the $(l+1)$-th text encoder layer. Final outputs of the visual and text encoders, $\mathbf{x}^{(L)}$ and $\mathbf{y}^{(L)}$, are denoted as follows:
\begin{equation}
\begin{split}
\mathbf{x}^{(L)} &= \left [e_0; E_x^{(L)};\{e_i\}_{i=1}^{K}\right], \\
\mathbf{y}^{(L)} &= \left[s_0; E_y^{(L)};\{s_i\}_{i=1}^{K}\right], \\
\end{split}
\end{equation}
\begin{equation}
\begin{split}
v_i = \mathbf{P}_{v} \cdot e_i, \\
t_i = \mathbf{P}_{t} \cdot s_i, \\
\label{eq:proj}
\end{split}
\end{equation}
where $L$ is the number of layers, $e_i, s_i$ are the $i$-th visual and text prompt feature, produced by their respective encoders. $\mathbf{P}_v$ and $\mathbf{P}_t$ are the pre-trained projection matrix that projects $e_i, s_i$ to $v_i, t_i$.

\subsection{Pairwise scoring function}
As shown in \Cref{fig:2}, for $K$ pairs of prompts, we compute $K$ logits based on cosine similarity given a single image $x$ and class label $y$. Given $x$ and $y$, we define the similarity between them as \Cref{eq:similarity}. By averaging the logits, we yield the same effect as an ensemble of $K$ independent models that have separate perspectives about image and text. 
\begin{equation} \text{sim}(x, y) = \frac{1}{K}\sum_{i=1}^{K} \frac{v_i\cdot t_i}{|v_i||t_i|} \label{eq:similarity} \end{equation}
\begin{equation} p(y_k|x) = \frac{\exp(\text{sim}(x, y_k)/\tau)}{\sum_{j=1}^{C} \exp(\text{sim}(x, y_j)/\tau)}
\label{eq:prob}
\end{equation}

Using ensembled logits, we define probability distribution following \Cref{eq:prob}, where $\tau$ denotes the temperature hyperparameter of pre-trained CLIP.\\


\begin{table*}[ht]
\caption{\textbf{Comparison of CLIP, CoOp, CoCoOp, and Ours (RPO) in the base-to-new generalization setting.} We train our model with a subset of the classes (base classes) in a 16-shot setting and evaluate on the test set including base classes and new classes. H denotes the harmonic mean of base and novel performance.}
  \begin{subtable}{0.3\linewidth}
  \centering
  \caption{Average over 11 datasets}
  \begin{tabular}{l|c|c|c}
    \toprule
    Methods & Base & Novel & H\\
    \midrule
    CLIP    &   \textcolor{DarkGray}{69.34}   &   \textcolor{DarkGray}{74.22}   &   71.70 \\
    \midrule
    +LP    &   \textcolor{DarkGray}{81.80}   &   \textcolor{DarkGray}{69.17}   &   
    74.65 \\
    +CoOp    &   \textcolor{DarkGray}{82.69}   &   \textcolor{DarkGray}{63.22}   &   71.66 \\
    +CoCoOp  &   \textcolor{DarkGray}{80.47}   &   \textcolor{DarkGray}{71.69}   &   75.83 \\
    +RPO    &   \textcolor{DarkGray}{81.13}    &   \textcolor{DarkGray}{75.00}    &   \textbf{77.78} \\
    
    \bottomrule
    \end{tabular}
    \label{tab:1.a}
  \end{subtable}
  \hfill
  \begin{subtable}{0.3\linewidth}
  \centering
  \caption{ImageNet.}
  \begin{tabular}{l|c|c|c}
    \toprule
    Methods & Base & Novel & H\\
    \midrule
    CLIP    &   \textcolor{DarkGray}{72.43}   &   \textcolor{DarkGray}{68.14}   &   70.22 \\
    \midrule
    +LP    &   \textcolor{DarkGray}{73.13}   &   \textcolor{DarkGray}{57.10}   &   64.13 \\
    +CoOp    &   \textcolor{DarkGray}{76.47}   &   \textcolor{DarkGray}{67.88}   &   71.92 \\
    +CoCoOp  &   \textcolor{DarkGray}{75.98}   &   \textcolor{DarkGray}{70.43}   &   73.10 \\
    +RPO    &   \textcolor{DarkGray}{76.60}    &   \textcolor{DarkGray}{71.57}    &   \textbf{74.00} \\
    
    \bottomrule
  \end{tabular}
  \label{tab:1.b}
  \end{subtable}
  \hfill
  \begin{subtable}{0.3\linewidth}
  \centering
  \caption{Caltech101.}
  \begin{tabular}{l|c|c|c}
    \toprule
    Methods & Base & Novel & H\\
    \midrule
    CLIP    &   \textcolor{DarkGray}{96.84}   &   \textcolor{DarkGray}{94.00}   &   95.40 \\
    \midrule
    +LP    &   \textcolor{DarkGray}{98.03}   &   \textcolor{DarkGray}{93.50}   &   95.71 \\
    +CoOp    &   \textcolor{DarkGray}{98.00}   &   \textcolor{DarkGray}{89.81}   &   93.73 \\
    +CoCoOp  &   \textcolor{DarkGray}{97.96}   &   \textcolor{DarkGray}{93.81}   &   95.84 \\
    +RPO     &   \textcolor{DarkGray}{97.97}  &   \textcolor{DarkGray}{94.37}    &   \textbf{96.03}  \\
    \bottomrule
  \end{tabular}
  \label{tab:1.c}
  \end{subtable}
  \hfill
  \vspace{0.5em}
  \begin{subtable}{0.3\linewidth}
  \centering
  \caption{OxfordPets.}
  \begin{tabular}{l|c|c|c}
    \toprule
    Methods & Base & Novel & H\\
    \midrule
    CLIP    &   \textcolor{DarkGray}{91.17}   &   \textcolor{DarkGray}{97.26}   &   94.12 \\
    \midrule
    +LP    &   \textcolor{DarkGray}{94.87}   &   \textcolor{DarkGray}{92.50}   &   93.67 \\
    +CoOp    &   \textcolor{DarkGray}{93.67}   &   \textcolor{DarkGray}{95.29}   &   94.47 \\
    +CoCoOp  &   \textcolor{DarkGray}{95.20}   &   \textcolor{DarkGray}{97.69}   &   \textbf{96.43} \\
    +RPO    &   \textcolor{DarkGray}{94.63}   &   \textcolor{DarkGray}{97.50}    &   96.05  \\
    \bottomrule
  \end{tabular}
  \label{tab:1.d}
  \end{subtable}
  \hfill
  \begin{subtable}{0.3\linewidth}
  \centering
  \caption{StanfordCars.}
  \begin{tabular}{l|c|c|c}
    \toprule
    Methods & Base & Novel & H\\
    \midrule
    CLIP    &   \textcolor{DarkGray}{63.37}   &   \textcolor{DarkGray}{74.89}   &   68.65 \\
    \midrule
    +LP    &   \textcolor{DarkGray}{78.60}   &   \textcolor{DarkGray}{65.50}   &   71.45 \\
    +CoOp    &   \textcolor{DarkGray}{78.12}   &   \textcolor{DarkGray}{60.40}   &   68.13 \\
    +CoCoOp  &   \textcolor{DarkGray}{70.49}  &   \textcolor{DarkGray}{73.59}   &   72.01 \\
    +RPO    &   \textcolor{DarkGray}{73.87}   &  \textcolor{DarkGray}{75.53}    &  \textbf{74.69}\\
    \bottomrule
  \end{tabular}
  \label{tab:1.e}
  \end{subtable}
  \hfill
  \begin{subtable}{0.3\linewidth}
  \centering
  \caption{Flowers102.}
  \begin{tabular}{l|c|c|c}
    \toprule
    Methods & Base & Novel & H\\
    \midrule
    CLIP    &   \textcolor{DarkGray}{72.08}   &   \textcolor{DarkGray}{77.08}   &   74.83 \\
    \midrule
    +LP    &   \textcolor{DarkGray}{97.87}   &   \textcolor{DarkGray}{65.87}   &   78.74 \\
    +CoOp    &   \textcolor{DarkGray}{97.60}   &  \textcolor{DarkGray}{59.67}   &   74.06 \\
    +CoCoOp  &   \textcolor{DarkGray}{94.87}   &   \textcolor{DarkGray}{71.75}   &   81.71 \\
    +RPO    &   \textcolor{DarkGray}{94.13}   &   \textcolor{DarkGray}{76.67}    &   \textbf{84.50}\\
    \bottomrule
  \end{tabular}
  \label{tab:1.f}
  \end{subtable}
  \hfill
  \vspace{0.5em}
  \begin{subtable}{0.3\linewidth}
  \centering
  \caption{Food101.}
  \begin{tabular}{l|c|c|c}
    \toprule
    Methods & Base & Novel & H\\
    \midrule
    CLIP    &   \textcolor{DarkGray}{90.10}   &   \textcolor{DarkGray}{91.22}   &   90.66 \\
    \midrule
    +LP    &   \textcolor{DarkGray}{88.30}   &   \textcolor{DarkGray}{88.03}   &   88.17 \\
    +CoOp    &   \textcolor{DarkGray}{88.33}   &   \textcolor{DarkGray}{82.26}   &   85.19 \\
    +CoCoOp  &   \textcolor{DarkGray}{90.70}   &   \textcolor{DarkGray}{91.29}   &   \textbf{90.99} \\
   +RPO     &  \textcolor{DarkGray}{90.33}   &   \textcolor{DarkGray}{90.83}   &  90.58   \\
    \bottomrule
  \end{tabular}
  \label{tab:1.g}
  \end{subtable}
  \hfill
  \begin{subtable}{0.3\linewidth}
  \centering
  \caption{FGVCAircraft.}
  \begin{tabular}{l|c|c|c}
    \toprule
    Methods & Base & Novel & H\\
    \midrule
    CLIP    &   \textcolor{DarkGray}{27.19}   &   \textcolor{DarkGray}{36.29}   &   31.09 \\
    \midrule
    +LP    &   \textcolor{DarkGray}{41.37}   &   \textcolor{DarkGray}{31.13}   &  35.53  \\
    +CoOp    &   \textcolor{DarkGray}{40.44}   &   \textcolor{DarkGray}{22.30}   &   28.75 \\
    +CoCoOp  &   \textcolor{DarkGray}{33.41}   &   \textcolor{DarkGray}{23.71}   &   27.74 \\
   +RPO     &   \textcolor{DarkGray}{37.33}  &    \textcolor{DarkGray}{34.20}   &  \textbf{35.70}   \\
    \bottomrule
  \end{tabular}
  \label{tab:1.h}
  \end{subtable}
  \hfill
  \begin{subtable}{0.3\linewidth}
  \centering
  \caption{SUN397.}
  \begin{tabular}{l|c|c|c}
    \toprule
    Methods & Base & Novel & H\\
    \midrule
    CLIP    &   \textcolor{DarkGray}{69.36}   &   \textcolor{DarkGray}{75.35}   &   72.23 \\
    \midrule
    +LP    &   \textcolor{DarkGray}{79.47}   &   \textcolor{DarkGray}{69.73}   &   74.28 \\
    +CoOp    &   \textcolor{DarkGray}{80.60}   &   \textcolor{DarkGray}{65.89}   &   72.51 \\
    +CoCoOp  &   \textcolor{DarkGray}{79.74}   &   \textcolor{DarkGray}{76.86}   &   78.27 \\
    +RPO    &    \textcolor{DarkGray}{80.60}   &  \textcolor{DarkGray}{77.80}     &  \textbf{79.18} \\
    \bottomrule
  \end{tabular}
  \label{tab:1.i}
  \end{subtable}
  \hfill
  \vspace{0.5em}
  \begin{subtable}{0.3\linewidth}
  \centering
  \caption{DTD.}
  \begin{tabular}{l|c|c|c}
    \toprule
    Methods & Base & Novel & H\\
    \midrule
    CLIP    &   \textcolor{DarkGray}{53.24}   &   \textcolor{DarkGray}{59.90}   &   56.37 \\
    \midrule
    +LP    &   \textcolor{DarkGray}{80.63}   &   \textcolor{DarkGray}{55.97}   &   66.07 \\
    +CoOp    &   \textcolor{DarkGray}{79.44}   &   \textcolor{DarkGray}{41.18}   &   54.24 \\
    +CoCoOp  &   \textcolor{DarkGray}{77.01}   &   \textcolor{DarkGray}{56.00}   &   64.85 \\
    +RPO     &   \textcolor{DarkGray}{76.70}    &   \textcolor{DarkGray}{62.13}    &  \textbf{68.61}  \\
    \bottomrule
  \end{tabular}
  \label{tab:1.j}
  \end{subtable}
  \hfill
  \begin{subtable}{0.3\linewidth}
  \centering
  \caption{EuroSAT.}
  \begin{tabular}{l|c|c|c}
    \toprule
    Methods & Base & Novel & H\\
    \midrule
    CLIP    &   \textcolor{DarkGray}{56.48}   &   \textcolor{DarkGray}{64.05}   &   60.03 \\
    \midrule
    +LP    &   \textcolor{DarkGray}{82.30}   &   \textcolor{DarkGray}{68.00}   &  74.47  \\
    +CoOp    &   \textcolor{DarkGray}{92.19}   &   \textcolor{DarkGray}{54.74}   &   68.69 \\
    +CoCoOp  &   \textcolor{DarkGray}{87.49}   &   \textcolor{DarkGray}{60.04}   &   71.21 \\
    +RPO     &  \textcolor{DarkGray}{86.63}   &  \textcolor{DarkGray}{68.97}   &   \textbf{76.79}  \\
    \bottomrule
  \end{tabular}
  \label{tab:1.k}
  \end{subtable}
  \hfill
  \begin{subtable}{0.3\linewidth}
  \centering
  \caption{UCF101.}
  \begin{tabular}{l|c|c|c}
    \toprule
    Methods & Base & Novel & H\\
    \midrule
    CLIP    &   \textcolor{DarkGray}{70.53}   &  \textcolor{DarkGray}{77.50}   &   73.85 \\
    \midrule
    +LP    &   \textcolor{DarkGray}{85.27}   &   \textcolor{DarkGray}{73.53}   &  78.97  \\
    +CoOp    &   \textcolor{DarkGray}{84.69}   &   \textcolor{DarkGray}{56.05}   &   67.46 \\
    +CoCoOp  &   \textcolor{DarkGray}{82.33}   &   \textcolor{DarkGray}{73.45}   &   77.64 \\
    +RPO    &   \textcolor{DarkGray}{83.67}    &  \textcolor{DarkGray}{75.43}    &   \textbf{79.34}\\
    \bottomrule
  \end{tabular}
  \label{tab:1.l}
  \end{subtable}
  \hfill
  \vspace{0.5em}
\label{tab:1}
\end{table*} 


\begin{table*}[ht]
    \centering
    \caption{\textbf{Comparison of RPO, CoCoOp, CoOp and manual prompt in domain generalization.} RPO learns from ImageNet (16 images per class) and is evaluated by 4 datasets with distribution shift and ImageNet itself. RPO performs better on 4 out of 5 datasets compared to CoCoOp.}
    \begin{tabular}{lcccccc}
        \toprule
        \textbf{} & & Source & \multicolumn{4}{c}{Target}  \\
          \cmidrule(r){3-3}\cmidrule(r){4-7}
          &{Learnable?} & {ImageNet} & {ImageNetV2} & {ImageNet-Sketch} & {ImageNet-A} & {ImageNet-R}\\
          \cmidrule(r){1-7}
        CLIP &  &  66.73 & 60.83 & 46.15 & 47.77 & 73.96\\
        +CoOp & \checkmark &  71.51 & 64.20 & 47.99 & 49.71 & 75.21\\
        +CoCoOp & \checkmark & 71.02 & 64.07 & 48.75 & \textbf{50.63} & 76.18\\
        \midrule
        +RPO & \checkmark &  \textbf{71.67} & \textbf{65.13} & \textbf{49.27} & \ 50.13 & \textbf{76.57} \\
    \bottomrule
  \end{tabular}
  \label{tab:2}
\end{table*}

\section{Experiments}
\label{sec:experiments}

Following CoCoOp~\cite{cocoop}, we evaluate our model, RPO, in two experimental settings, 1) Base-to-new generalization, which aims to demonstrate generalization to the label-shift, where labels are divided into base and novel classes, and 2) domain generalization, which aims to show generalization to the domain shift, especially for out-of-distribution data. We also conduct extensive analyses to explore RPO's capability to reduce model variance and improve generalization while maintaining parameter efficiency and computational efficiency.\\

\noindent\textbf{Datasets}
We evaluate RPO in label-shift on 11 image recognition datasets used in CoOp~\cite{coop} and CoCoOp~\cite{cocoop}. Specifically, we use ImageNet~\cite{imagenet}, Caltech101~\cite{caltech101}, OxfordPets~\cite{oxfordpets}, StanfordCars~\cite{stanfordcars}, Flowers102~\cite{flowers102}, Food101~\cite{food101}, FGVCAircraft~\cite{fgvcaircraft}, SUN397~\cite{sun397}, DTD~\cite{dtd}, EuroSAT~\cite{eurosat}, and UCF101~\cite{ucf101}. We also conduct experiments to evaluate the domain generalization ability of RPO with ImageNet~\cite{imagenet} as the source dataset and its distinct-domain variants ImageNetV2~\cite{imagenetv2}, ImageNet-Sketch~\cite{imagenetsketch}, ImageNet-A~\cite{imageneta}, and ImageNet-R~\cite{imagenetr} as the target datasets.\\

\noindent\textbf{Baselines}
We set our baseline as CoCoOp~\cite{cocoop} for two experiments: base-to-new generalization and domain generalization. We compare RPO with zero-shot CLIP~\cite{clip} based on manually chosen prompt templates for each dataset and CoOp~\cite{coop} which optimizes learnable context vectors. We also take into account the Linear-probing (LP CLIP) in our analysis. This approach involves incorporating an extra trainable linear layer on the existing CLIP image encoder. In contrast to the typical Linear-probing method, which solely relies on the CLIP image encoder and a trainable linear classifier, we additionaly utilize CLIP text embeddings which encode the classnames as a classifier weights to evaluate LP CLIP on base-to-new generalization setting. RPO shows better generalization and robustness compared to CoCoOp with fewer parameters and computational expenses, as shown in \Cref{tab:1} and \Cref{tab:2}.\\

\noindent\textbf{Training details}
In all the experiments, we use ViT-B/16 CLIP, a CLIP with vision transformer backbone, as our base model. We set the number of prompt pairs $K$ as $24$ for fair comparison with CoCoOp regarding the number of parameters. 
The SGD optimizer is used with batch size 4. For base-to-new generalization, RPO is trained for 15 epochs with a learning rate of 0.01. 
For domain generalization, we trained our model for 15 epochs with a learning rate of 0.005.

\subsection{Base-to-new generalization}
For each dataset, we split classes into two groups, base and novel, by the alphabetical order of labels.
The training dataset consists of 16 images per class of the base classes at random. Models are trained by this few-shot sampled data depending on 3 random seeds (1, 2, and 3) as \cite{cocoop}, and we report the averaged results in the \Cref{tab:1}. We evaluate accuracy on test data corresponding to both the base and novel classes and use their harmonic mean as the final evaluation metric. \\

\noindent\textbf{Comparison with CoCoOp}
RPO outperforms CoCoOp on 9 out of 11 image recognition benchmarks, while simultaneously addressing the computational cost associated with CoCoOp's instance-conditional design. See \Cref{analysis} for more discussions about the computational efficiency. \Cref{tab:1} shows that our method shows better generalization to label shift in most benchmarks. Out of 11 datasets, RPO achieved better accuracy in 8 of base and 9 of novel, compared with CoCoOp. In average over 11 datasets, the gap between the accuracy on base classes and novel classes decreased, indicating better base-to-new generalization. It is worth mentioning that the averaged novel accuracy of RPO surpasses the zero-shot CLIP and also outperforms zero-shot CLIP on 7 out of 11 benchmarks. It supports that RPO is a generalizable adaptation method for label shift. Although RPO brings slightly lower performance in OxfordPets and Food101 compared to CoCoOp, the result shows an overall improvement in both base classes and novel classes.\\

\noindent\textbf{Comparison with CoOp}
RPO and CoOp share similar architectures in that both of them introduce learnable prompts only into the input space. Despite the architectural similarity, RPO results in higher novel accuracy on all datasets compared to CoOp. As shown in \Cref{tab:1}, RPO improves the novel accuracy of CoOp by 11.8\% on average, which far outweighs the 1.5\% drop in base accuracy. It demonstrates the fact that read-only prompts implemented by masked attention result in a better base to new generalization in the context of vision-language model adaptation.\\

\noindent\textbf{Comparison with LP}
Additionally, Linear-Probing (LP CLIP), introduced in \Cref{fig:0}, can be considered a comparable baseline for base-to-new generalization. Despite not outperforming on every benchmark, LP CLIP's competitive performance and its relatively small performance variation implies preventing internal representation shift is beneficial for robust fine-tuning in data-deficient settings. Despite the commonality that both LP CLIP and RPO do not shift the internal representation, RPO exhibits superior generalization performance across 11 datasets. This observation aligns with previous works \cite{vistune, low-resource-sp, prefix} in that the prompt tuning outperforms the conventional fine-tuning methods in low-data scenarios.\\

\subsection{Domain generalization}
 By measuring the generalization ability of the model on out-of-distribution data, we can verify how robust our learned prompts are to domain shift. In this section, we evaluate RPO's domain generalization performance. We first train RPO with all classes of ImageNet on the 16-shot setting and then evaluate accuracy on out-of-distribution datasets (ImageNetV2~\cite{imagenetv2}, ImageNet-Sketch~\cite{imagenetsketch}, ImageNet-A~\cite{imageneta}, and ImageNet-R~\cite{imagenetr}). As shown in \Cref{tab:2}, compared to CoCoOp, RPO achieves better generalization performance on the four datasets, except for ImageNet-A. This shows that RPO is more robust to out-of-distribution.\\

\begin{figure}[h]
\centering
{\includegraphics[width=8.5cm,height=19.5cm]{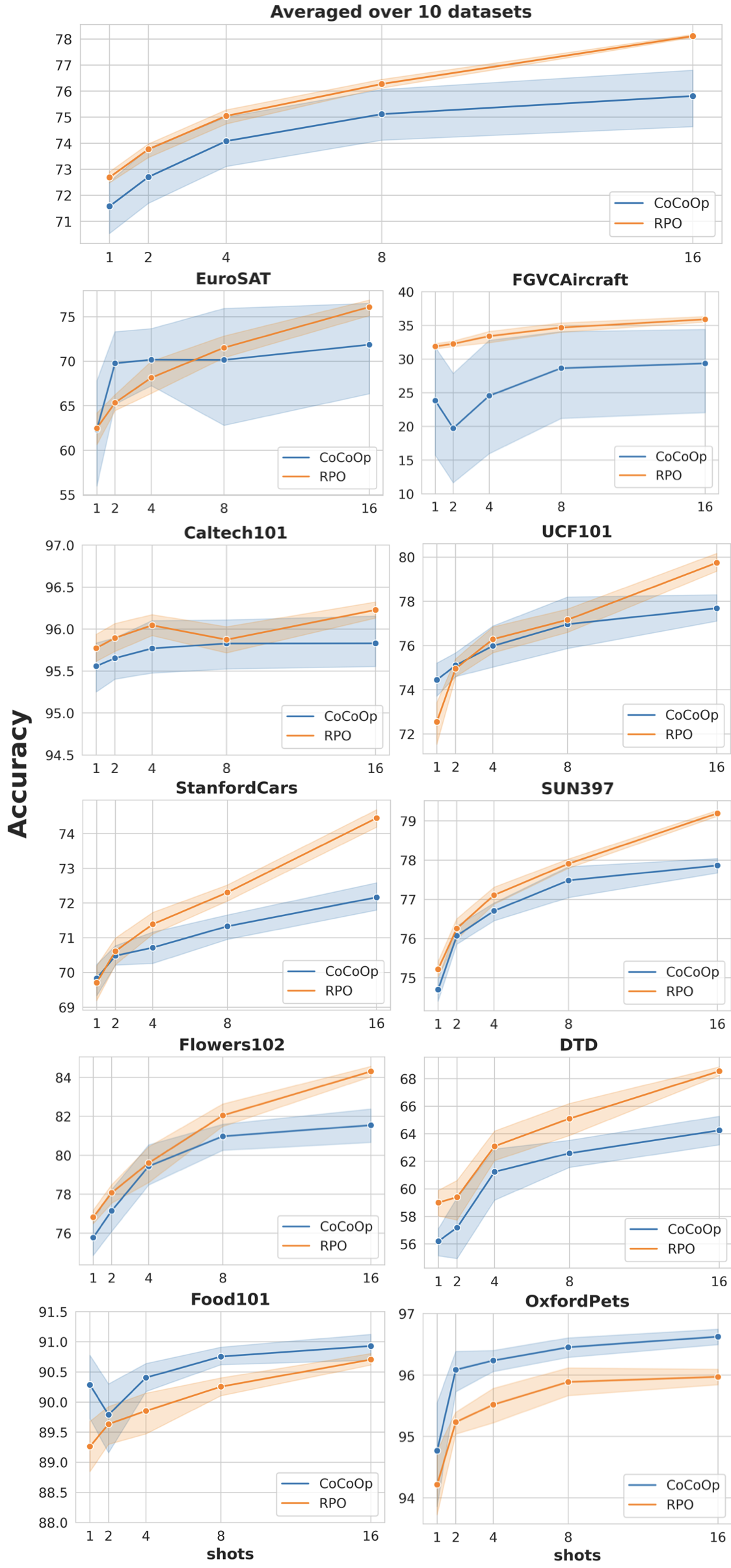}} 
\caption{\textbf{Variance and generalization of RPO compared with CoCoOp.} RPO is more generalizable and robust than CoCoOp in the perspective of base to new generalization and lower performance variance.}
\label{fig:5}
\end{figure}

\subsection{Analysis}
\label{analysis}

\begin{table}[h]
\caption{\textbf{Ablation result averaged over 11 datasets.} }
  \centering
  \begin{tabular}{l|c|c|c}
    \toprule
    Methods & Base & Novel & H\\
    \midrule
    RPO w.o mask/init & \textcolor{DarkGray}{78.63} & \textcolor{DarkGray}{69.56} &  73.29\\
    RPO w.o mask & \textcolor{DarkGray}{78.55} & \textcolor{DarkGray}{71.34} & 74.59 \\
    RPO w.o init & \textcolor{DarkGray}{82.00} & \textcolor{DarkGray}{72.94} & 76.82 \\
    RPO    &   \textcolor{DarkGray}{81.13}    &   \textcolor{DarkGray}{75.00}  &   \textbf{77.78} \\
    \bottomrule
    \end{tabular}
\label{tab:3}
\end{table}

\noindent\textbf{Ablation on masked attention and ST-initialization}
We conduct an ablation study to measure the effect of the read-only mechanism and ST-initialization. We evaluate 3 variants of RPO (without an attention mask, without ST-initialization, and without both) on the base to new generalization setting using 11 image recognition datasets. We report averaged accuracy in \Cref{tab:3} to demonstrate that the combination of masked attention and ST-initialization leads to better generalization performance. More detailed ablation studies with each dataset is presented in the supplement.\\

\noindent\textbf{Analysis on model variance and extreme few-shot setting}
If the training samples for adaptation are limited (\eg, less than 16 samples per class) in real-world scenarios, the model variance has a higher chance of getting large.
 To show the advantage of RPO in alleviating model variance as well as improving base-to-new generalization in extreme few-shot settings (training samples less than 16 per class), we train the model with 10 random seeds for 10 benchmarks~\cite{caltech101, oxfordpets, stanfordcars, flowers102, food101, fgvcaircraft, sun397, dtd, eurosat, ucf101} on 1, 2, 4, 8, and 16-shot settings. We set the number of prompts $K$ as 4 in this analysis.
Then, we compute the harmonic mean of base and novel accuracy for each of the 10 random seeds and visualize their variance in \Cref{fig:5}.
As shown in the \Cref{fig:5}, RPO shows remarkably lower variance compared to CoCoOp on average, which supports the effectiveness of the read-only mechanism.
Especially, in a 16-shot setting, RPO reduced the variance by 94\% on average compared to CoCoOp, which is demonstrated in the \Cref{tab:4}. This demonstrates that RPO stabilizes performance variance on 10 benchmarks, including EuroSAT and FGVCAircraft benchmarks, where CoCoOp exhibits extremely high variance.
Also, RPO shows superior base to new generalization. As demonstrated in \Cref{fig:5}, RPO results in more than 1\% higher harmonic mean score compared to CoCoOp  on every shot (1, 2, 4, 8, and 16).
We conjecture that the lower variance and the better generalization comes from the characteristics of RPO that prevents the internal representation shift of pre-trained model.\\



\begin{table*}[t]
\small
    \centering
    \setlength{\tabcolsep}{2pt}
    \caption{\textbf{Analysis of RPO on extreme few shot settings.} We report RPO's averaged base accuracy, novel accuracy, and their harmonic mean on 10 benchmark datasets. RPO consistently outperforms CoCoOp on 1, 2, 4, and 8 shot setting evaluated by harmonic mean.}
    \begin{tabular}{lcccccccccc}
        \toprule
        & \multicolumn{2}{c}{1 shot} & \multicolumn{2}{c}{2 shot}& \multicolumn{2}{c}{4 shot} & \multicolumn{2}{c}{8 shot} & \multicolumn{2}{c}{16 shot} \\
          \cmidrule(r){2-3}\cmidrule(r){4-5}\cmidrule(r){6-7}\cmidrule(r){8-9}\cmidrule(r){10-11}
          &{CoCoOp} & {RPO} & {CoCoOP} & {RPO} & {CoCoOp} & {RPO} & {CoCoOp} & {RPO} & {CoCoOp} & {RPO}\\
          \cmidrule(r){1-11}
        Base  & 71.45$\pm$1.58 & \textbf{71.69}$\pm$\textbf{0.30}  & \textbf{73.93}$\pm$1.26 & 73.82$\pm$\textbf{0.57} & 76.50$\pm$0.96 & \textbf{77.18}$\pm$\textbf{0.71}& 78.46$\pm$1.02 & \textbf{79.66}$\pm$\textbf{0.36} & 80.57$\pm$0.60 & \textbf{81.31}$\pm$\textbf{0.30} \\
        Novel & 72.47$\pm$2.00 & \textbf{73.82}$\pm$\textbf{0.73}& 71.91$\pm$2.25 & \textbf{73.83}$\pm$\textbf{0.64} & 72.50$\pm$2.06 & \textbf{73.43}$\pm$\textbf{0.67} &  72.78$\pm$2.10 & \textbf{73.66}$\pm$\textbf{0.50} & 72.51$\pm$2.19 & \textbf{75.47}$\pm$\textbf{0.25}\\
        \midrule
        H.M &71.78$\pm$1.80& \textbf{72.69}$\pm$\textbf{0.37} & 72.70$\pm$1.80 & \textbf{73.77}$\pm$\textbf{0.45} & 74.08$\pm$1.63 & \textbf{75.05}$\pm$\textbf{0.45} &  75.12$\pm$1.74 & \textbf{76.27}$\pm$\textbf{0.28} & 75.81$\pm$1.77 & \textbf{78.11}$\pm$\textbf{0.10}\\
    \bottomrule
  \end{tabular}
  \label{tab:4}
\end{table*}

\begin{table}[h]
\caption{\textbf{Generalizability of uni-modal RPO.}}
  \centering
  \begin{tabular}{l|c|c|c}
    \toprule
    Methods & Base & Novel & H\\
    \midrule
    CoOp  &   \textcolor{DarkGray}{82.69}   &   \textcolor{DarkGray}{63.22}   &   71.66 \\
    CoCoOp  &   \textcolor{DarkGray}{80.47}   &   \textcolor{DarkGray}{71.69}   &   75.83 \\
    \midrule
    text-RPO & \textcolor{DarkGray}{79.54} & \textcolor{DarkGray}{74.84} & 77.01 \\
    RPO    &   \textcolor{DarkGray}{81.13}    &   \textcolor{DarkGray}{75.00}    &   \textbf{77.78} \\
    \bottomrule
    \end{tabular}
\label{tab:5}
\end{table}


\noindent\textbf{RPO with uni-modal prompts}
For a better understand of RPO in each modality, we experiment with RPO with only text prompts (text-RPO) with little modification to the pairwise scoring function. Text-RPO and CoOp differ in the point that RPO's prompts do not affect the internal representation of the pre-trained model but CoOp's prompts do. As shown in \Cref{tab:5}, uni-RPO still achieves competitive performance compared to CoCoOp with a 0.8\% drop compared to RPO, which again demonstrates the effectiveness of the read-only mechanism.\\

\begin{figure}[h]
\centering
{\includegraphics[width=8cm,height=4cm]{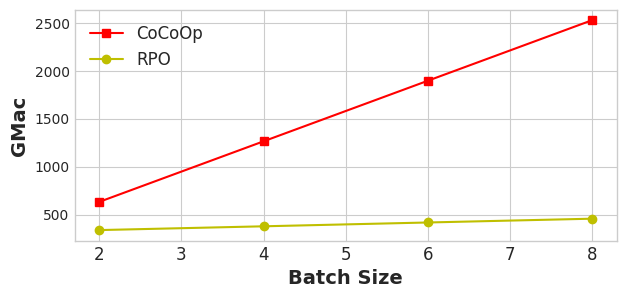}} 
\caption{\textbf{Computational Cost of CoCoOp an RPO.}}
\label{fig:6}
\end{figure}

\noindent\textbf{Computational efficiency}
It is worth highlighting that RPO surpasses CoCoOp in both generalization performance and computation efficiency. Note that CoCoOp employs image conditional prompts depending on the input image, resulting in a significant increase in computational overhead. Considering self-attention, roughly speaking, CoCoOp's computational complexity of $O(BCN_t^2 + BN_v^2)$, where $B$, $C$, $N_t$, and $N_v$ represent the batch size, the number of classes in the dataset, the length of the text tokens, and the length of the image patches, respectively. On the other hand, RPO achieves better generalization performance when compared to CoCoOp, while maintaining the same computational complexity as CoOp, which is roughly speaking $O(CN_t^2 + BN_v^2)$ regarding self-attention. 
As shown in \Cref{fig:6}, we measure computational overhead (GMac) for inference with respect to the batch size. When examining the rate of increase in GMac with respect to the increase in batch size, the rate of increase exhibited by CoCoOp is significantly greater than that of RPO.

\section{Conclusion}
\label{sec:conclusion}


The emergence of large-scale, pre-trained models like CLIP~\cite{clip}, ALIGN~\cite{ALIGN}, and FILIP~\cite{FILIP} has made it increasingly important to efficiently adapt them to downstream tasks in parameter-efficient manner. Fine-tuning the entire model can be resource-intensive and may damage the well-defined model representations learned during pre-training. In perspective of the parameter efficiency, prompt learning is a promising approach to avoid these issues, but existing methods still end up shifting the representation of data tokens through attention mechanism~\cite{coop, cocoop, unsupervised}, which is an unstable adaptation strategy especially in data-deficient settings such as few-shot learning.

To address these challenges, we propose a novel approach that utilizes read-only prompts to prevent internal representation shift in the backbone model, resulting in better generalization and robustness. Our approach also employs learnable prompts on both the visual and text encoder, and we initialize them to special tokens like \texttt{[CLS]} and \texttt{[EOS]} for better convergence. Our extensive experiments demonstrate that our approach outperforms other methods in base-to-new generalization and domain generalization with remarkably lower variance.

However, despite the significant potential of this approach, it remains an under-explored area. Further research is needed to fully understand the efficiency and effectiveness of this method compared to other adaptation strategies. Nevertheless, our approach offers a promising direction for a generalizable and robust adaptation of pre-trained models in resource-limited settings.

\section*{Acknowledgments}
This research was in part supported by the MSIT (Ministry of Science and ICT), Korea, under the ICT Creative Consilience program (IITP-2023-2020-0-01819) supervised by the IITP (Institute for Information \& communications Technology Planning \& Evaluation); the National Research Foundation of Korea (NRF) grant funded by the Korea government (MSIT) (NRF-2023R1A2C2005373); and KakaoBrain corporation.
{\small
\bibliographystyle{ieee_fullname}
\bibliography{egbib}
}
\end{document}